\pdfoutput=1

\documentclass[11pt]{article}

\usepackage{acl}

\usepackage{times}
\usepackage{latexsym}
\usepackage{graphicx}
\usepackage{caption}
\usepackage{subcaption}
\usepackage{soul}
\usepackage{algorithmicx}
\usepackage{algpseudocode}
\usepackage{algorithm}
\usepackage{amssymb}
\usepackage{dblfloatfix}
\usepackage{mdframed}
\usepackage{makecell}

\usepackage[T1]{fontenc}

\usepackage[utf8]{inputenc}

\usepackage{microtype}

\usepackage{inconsolata}

\algrenewcommand\algorithmicrequire{\textbf{Input}}
\renewcommand*\Call[2]{\textproc{#1}(#2)}

\title{Policies and Evaluation for Online Meeting Summarization}

\author{Felix Schneider \and Marco Turchi\\
  Zoom Video Communications \\
  \texttt{firstname.lastname@zoom.us}\\\And
  Alex Waibel \\
  Karlsruhe Intstitute of Technology\\
  \texttt{waibel@kit.edu}
  }

\begin{document}
\maketitle
\begin{abstract}

With more and more meetings moving to a digital domain, meeting summarization has recently gained interest in both academic and commercial research. However, prior academic research focuses on meeting summarization as an offline task, performed after the meeting concludes. In this paper, we perform the first systematic study of \emph{online} meeting summarization. For this purpose, we propose several policies for conducting online summarization. We discuss the unique challenges of this task compared to the offline setting and define novel metrics to evaluate latency and partial summary quality. The experiments on the AutoMin dataset show that 1) online models can produce strong summaries, 2) our metrics allow a detailed analysis of different systems' quality-latency trade-off, also taking into account intermediate outputs and 3) adaptive policies perform better than fixed scheduled ones. These findings provide a starting point for the wider research community to explore this important task.

\end{abstract}

\section{Introduction}
\label{sec:intro}


In recent years, in particular due to the COVID-19 pandemic, the way we work has changed, moving more activities into a digital space. This drastically increased the number of meetings taking place remotely via video conferencing. As a result, activities relating to physical meetings, such as note-taking, transcription, and summarization are also becoming digitized. Among these activities, automatic summarization has received particular attention from the research and industrial community, resulting in automatic summarization systems that can assist human participants in creating meaningful meeting notes or summaries. Neural network models and particularly Large Language Models (LLMs) have produced impressive results in this area in recent months, moving meeting summarization from being purely a research topic into practical usefulness. This observation was validated in the evaluation of the 2023 Workshop on Automatic Minuting (AutoMin, \citealt{ghosal-etal-2023-overview}), where humans rated the output of several automatic systems higher than the human-written reference summary.

However, existing research on meeting summarization as an offline task, performed after the meeting is completed. The summary as an artifact is only considered useful as a retrospective of the meeting's content. We lack any study of meeting summarization as an online task, producing summaries as the meeting is still in progress. Online summaries can be useful for catching up with participants coming in late, they can provide live feedback on the coverage of an agenda, and assist in note-taking.

Producing summaries online introduces unique challenges: Summaries must be produced based on incomplete information and possibly updated to reflect new information. Most of the time, users will see these intermediate summaries, making it all the more important to study their quality. This goal must be balanced against the demand of the user to see summarized content as soon as possible and for the output not to fluctuate. More specifically, the three axes of quality for online summarization are: \textbf{Quality}, usually defined by some variation of adequacy (coverage and accurate representation of all important content), fluency (readability of the text), and relevance (covering only important content, \citealt{fabbri2021summeval,ghosal-etal-2023-overview}). Unique to online summarization is that we must evaluate not only the final output but also the intermediate summaries. 
\textbf{Latency}, defined as the time between a piece of content\footnote{Following \citet{fabbri2021summeval}, we refer to information from the source as ``content'' rather than syntactic terms like ``sentence'' or ``word'', which need not be copied to the summary verbatim or semantic terms such as ``fact'' because a summary can contain information about opinions or hypotheses.} being generated in the output and that same content appearing in the source. It is desirable that relevant content from the source appears in the summary as soon as possible. \textbf{Rewriting}, the changing of previous output in order to correct mistakes or update information. This adds cognitive strain and should be used only when necessary. There is a lack of appropriate metrics to measure quality along these three axes as well as suitable evaluation data.


In this paper, we conduct the first systematic study of applying meeting summarization systems in an online fashion. For this purpose, we present several policies to address the task of online summarization. These policies assume the presence of an offline summarizer and build on top of it different criteria to decide when to read tokens from an input stream or write tokens to an output stream. We evaluate these policies on the AutoMin 2023 test set, according to the three criteria above: Quality (including intermediate summaries), latency, and rewriting, using existing and novel purpose-built metrics. Using our metrics, we examine the different quality-latency trade-offs of the different policies. Our evaluation shows that while the online systems cannot yet reach the same level of quality as offline ones, they nevertheless produce strong summaries (rated 4 out of 5 or higher by humans).

In summary, our contributions are as follows:

\begin{itemize}
    \item We conduct the first study of online meeting summarization, proposing and testing several policies based on established strong summarization systems.
    \item We introduce novel metrics for intermediate quality and output latency, which we use to compare the systems. We validate these metrics with a human evaluation.
    \item Our evaluations show that our proposed systems are able to generate high-quality summaries of meetings in an online setting.
\end{itemize}

\section{Related Work}


Meeting summarization has been studied for decades, with most pre-neural network approaches focusing on extractive summarization \citep{waibel1998meeting,zechner2002automatic,tur2008calo,garg2009clusterrank,tixier2017combining}. While extractive summaries can be perfectly adequate for some domains like news \citep{zhong2020extractive}, humans tend to prefer abstractive summaries \citep{murray2010generating,zhang2023benchmarking} and most current research is focused on generative (neural) abstractive summarization models \citep{zhu2020hierarchical,fabbri2021convosumm,zhang2022summn}.

Research into abstractive meeting summarization has largely focused on neural methods, with work largely progressing in two complementary directions, both attempting to cope with the challenge of long inputs: 1) extending the transformer \citep{vaswani2017attention} architecture to scale to longer inputs, usually using a sparse attention paradigm \citep{brown2020language,fabbri2021convosumm,zhong2022dialoglm} or a hierarchical representation approach \citep{zhu2020hierarchical}. And 2) divide-and-conquer approaches that perform a segmentation on the source and use conventional summarization systems on the segments, sometimes with an additional refinement step on the segment summaries \citep{zhang2022summn,asi2022end,liu2022dynamic}. The best summaries are currently obtained by Large Language Models (LLMs) producing abstractive summaries \citep{zhang2023benchmarking}. 

Online summarization has largely been studied in the context of stream summarization and video summarization \citep{sequiera2018overview}. In stream summarization, it is assumed that a large volume of content (usually tweets or news articles) is being generated all the time and the task is to summarize available content or sentiment on a given topic from this stream at a given time.
The primary concern of these systems is to produce a summary of past content on demand, not to respond to new content in a timely fashion. As a result, the concept of latency is not usually discussed in this context.
Typical approaches include performing document retrieval as a summary \citep{ge2016news} and constructing a language model from the content stream and generating a representative example as a summary \citep{olariu2014efficient}. Both require orders of magnitude more content than is available in a typical meeting and cannot be applied to meeting summaries.

Video summarization is the task of selecting representative frames or sequences from a video to form a summary \citep{apostolidis2021video}. While online systems exist \citep{lal2019online,zhao2014quasi}, they are very specific to video and exclusively produce extractive summaries. Moreover, there exists, to our knowledge, no study of latency (how soon after the occurrence is a summary available) in this task.


In summary, while there are systems that perform summarization in an online fashion, their approaches and evaluation schemes are not suitable for our task.



\section{Online Summarization}
\label{sec:online}

Adapting the paradigm from simultaneous machine translation \citep{gu2017learning}, we treat the summarization system as an agent, reading from a stream of input tokens and writing a stream of output tokens. The lengths of either of those streams are initially unknown. The behavior of the agent is determined by a \emph{policy}, which at any given time can choose between three different actions: \texttt{READ} content from the input stream, \texttt{WRITE} content to the output stream, or \texttt{ERASE} previously written content from the output stream. While in principle the system could read and individual tokens, in this work, we restrict ourselves to systems that read chunks of $C$ tokens at once and write whole summary sentences at once.



\subsection{Policies}

We propose several baseline policies for evaluation. Following recent findings from simultaneous machine translation \citep{papi-etal-2022-simultaneous}, we focus on parameter-free policies that are applied post-training to an offline-trained summarization system, controlling when and with what input the model is used. Such policies have the advantage of being largely agnostic to the backend summarizer and being quick to develop and deploy, easing their real-world adoption.

\paragraph{Length-Based Segmentation} This policy reads chunks one at a time from the input and summarizes them independently. This method is simple, has a configurable latency (the chunk size $C$), has no processing redundancy (every input token is processed exactly once), and does not rewrite its output. However, because the chunks are summarized independently, there is a risk of redundancy in the output.
We theorize that a summarization system will have the least redundancy and best coverage if each summarization unit (the span of input that is the basis of a summary shown to the user) is about a single topic and the boundaries between the units also represent boundaries between topics in the meeting. The boundaries selected by this policy are not likely to align with topic boundaries, which may hurt the final quality. This motivates the following two policies.

\paragraph{Model-Based Segmentation} This policy uses quality estimation to determine the ideal summarization unit size. The algorithm works as follows (pseudocode can be found in Appendix~\ref{app:pseudo}):

\begin{enumerate}
    \item Read one chunk. Summarize.
    \item Read another chunk. Summarize it together with the previous chunk(s).
    \item Repeat step 2 until the maximum input size is reached.
    \item Out of these summaries, write the one with the highest estimated quality.
    \item Discard all chunks that were used for that summary and restart at 1. This will cause the remaining chunks to be read again.
\end{enumerate}

The boundaries of the summarization units have a profound effect on the quality, and a fixed segmentation policy may not produce ideal results. This policy represents a pragmatic approach of simply trying different chunk sizes and showing only the summary that is judged best by a (reference-free) quality estimation model. In this experiment, we use model confidence as the scoring function, but other metrics, like cosine similarity to the source \cite{pham2023select} are also possible.

This policy will usually incur more latency than the length-based one because it always needs to completely fill the model input. We theorize that the adaptive segmentation will align more closely with topic boundaries, reducing redundancy in the output. Because this policy re-ingests chunks, it also has some processing redundancy. Like the length-based policy, its output is purely incremental, it does not revise.

\paragraph{Sliding Window} This policy takes a different approach to estimating the ideal unit size, by using the discrete model outputs directly. It works under the assumption that when adding more input, while keeping the previous output as a fixed prefix, at some point the model will add additional content to the output. For example, for an input of 10 turns, the model would write a one-sentence summary. When adding another 10 turns, and fixing the previous output as a prefix, the model would add a second sentence. The point when the model adds more output is taken to represent a boundary condition. We then rotate out the old input and output and continue. In detail, the policy works as follows (see also Appendix~\ref{app:pseudo} for pseudocode):

\begin{enumerate}
    \item The algorithm maintains a window of input chunks and a summary prefix. Both are initially empty.
    \item Read one chunk into the input window. Summarize, write the result. The summary becomes the prefix.
    \item Read another chunk into the window and summarize all input in the window. The summarizer is forced to begin with the prefix. If this produces new output, go to 4. If there is no new output (the summarizer ends the output immediately after the prefix), repeat 3.
    \item Write the new output, without the prefix. All chunks that were used for the old prefix rotate out of the input window. The new output becomes the prefix. Go to 3.
\end{enumerate}

This policy does not rely on external quality estimation, only using the model's outputs as a segmentation signal. It is also aware of its previous outputs, which could allow it to reduce redundancy compared to the independent policies. The lower bound for latency would be equal to the chunk-based policy if the model updates its output after every chunk. Because past input is kept in the buffer, this policy will incur at least a 2x processing redundancy.

\paragraph{Full Rewriting} If we have a summarization model with a sufficiently large input size (e.\,g.\ GPT-4), we can simply read chunks one at a time and summarize the whole input up to that point every time, erasing the previous output.
This policy is expected to have very high processing redundancy and a high degree of rewriting. The delay is expected to be the same as the length-based policy's because both policies update their summaries after every chunk. Among the proposed policies, it is also the only one able to effectively limit its total output length. The other policies' outputs will grow unbounded, roughly linear with the input size, which may not be desired. By dropping previously written content from the summary, this policy would be able to prevent the summary from growing indefinitely.

\paragraph{Fully Incremental} Similar to the rewriting policy above, after reading a chunk, this policy re-summarizes all input up to that point. However, its previous output is forced as a prefix, similar to the sliding window policy. This eliminates the problem of excessive rewriting, at the cost of the summary growing potentially unbounded again. This policy also requires a model with a long enough input size to accommodate the entire input.

\section{Data}

There are several meeting summarization datasets available, including AMI~\citep{kraaij2005ami}, ICSI~\citep{janin2003icsi}, QMSum~\citep{zhong2021qmsum} and AutoMin/Elitr~\citep{nedoluzhko-etal-2022-elitr,ghosal-etal-2023-overview}. AMI consists of staged meetings with very repetitive content, which makes it not ideal. This also rules out QMSum, which is mostly based on AMI. ICSI and AutoMin consist of real meetings of realistic length with varying numbers of participants. We decide to test on the more recent AutoMin, specifically the 2023 test set consisting of 12 meetings because the reference summaries are of high quality and we have a recent \citep{ghosal-etal-2023-overview} comprehensive study about how the automatic metrics correlate with human judgment. Since we focus on the text-to-text setting for this evaluation, audio is not required. Similar to \citet{schneider-turchi-2023-team}, we found it beneficial for summary quality to reverse the deidentification of the AutoMin data by replacing the \texttt{PERSON}, \texttt{ORGANIZATION} and \texttt{PROJECT} labels with random names, 4- and 3-letter acronyms, respectively. This conversion is reverted before calculating any metrics. There is no audio and thus no timing information available for AutoMin, so we assume unit duration for each dialog turn for calculating latency.

\section{Models}

Our policies can be applied to any offline summarization model. For this paper, we use two models: Bart-large~\citep{lewis2020bart}, fine-tuned with SamSum~\citep{gliwa2019samsum}\footnote{\href{https://huggingface.co/lidiya/bart-large-xsum-samsum}{\texttt{lidiya/bart-large-xsum-samsum}} on HuggingFace} and OpenAI's GPT-4~\citep{gpt4}, specifically \texttt{gpt-4-32k}, for the increased context length. Bart cannot be directly trained on the AutoMin training data, because most of the meetings are longer than its maximum context length of 1024 tokens. Fine-tuning on Samsum is the next best thing, as it is a high-quality dialog summarization dataset \citep{asi2022end}. GPT-4 is used without further customization. Wherever possible, we apply all policies to both models. However, Bart cannot be used with the full rewriting and fully incremental policies, because of its input length limitation, and GPT-4 cannot be used with the model-based policy using model confidence, because OpenAI's proprietary API does not report output probabilities. Prompts and hyperparameters can be found in Appendix~\ref{app:prompts}.

Wherever possible, we test the systems with chunk sizes 256, 512 and 1024, except for the model-based and sliding window algorithms, which are processing multiple chunks at once. In order to fit more than one chunk into Bart's input, we limit the chunk size to 512. GPT-4 has no such restriction, so we test with all of the above chunk sizes, as well as 2048.

We compare our systems to the best LLM and non-LLM models on this dataset from the AutoMin evaluation campaign \citep{ghosal-etal-2023-overview}: GPT-4, using an undisclosed method of segmentation and ``Zoom-long'' \citep{schneider-turchi-2023-team}, which is based on Bart using topic segmentation and finetuning with LLM output. The summaries from both systems are made available by the organizers. 

\section{Evaluation Metrics}

\subsection{Quality} 
We measure the final summary quality with ROUGE-1 F1~\citep{lin2004rouge}\footnote{We used the \href{https://pypi.org/project/py-rouge/}{Py-rouge} implementation.}, because \citet{ghosal-etal-2023-overview} showed this metric to have the best correlation with human judgment on this dataset.

\subsection{Latency} 
For measuring latency, we considered metrics from machine translation (length-adjusted average lagging, LAAL, by \citealt{papi2022over} is currently one of the most widely used ones). However, these metrics come with an implicit assumption that the hypothesis will be roughly the same length as the reference. In our evaluation, we encounter systems with widely varying hypothesis lengths, which distorts the existing latency metrics. This is evident most clearly when comparing two systems that operate on exactly the same schedule, such as length-based Bart 1024 and GPT-4 1024. These systems should have similar latency, but they have drastically different LAAL scores (112.5 vs. 67.7).

Recall our definition of latency from Section~\ref{sec:intro}: Latency is the time between a piece of content appearing in the output and its appearance in the source. Ideally, this should be established by an alignment between the output and the source, to identify which content from the output matches with which parts of the source. Producing such an alignment is a research topic in itself, so for this evaluation, we propose our own metric, \emph{Expected Latency} (EL), based on a simplifying assumption: That the summarization output contains all relevant content from the source that appeared since the last output. Given this simplification, calculating latency reduces to answering the question: ``Sampling a random point in the meeting, how long do I have to wait for the next summary?''.  This gives us an easily interpretable metric with which we can reason about the delay of different systems. Formally, EL is defined as: $EL = E[N(t) - t]$, where $t$ is a point in time from a uniform distribution over the input and $N(t)$ is the timestamp of the next WRITE event after time $t$. In practice, this is calculated as:

\begin{equation}
EL = \frac{1}{T} \sum_{t \in S \cup \{0\}} \frac{(N(t) - t)^2}{2}    
\end{equation}

Where $T$ is the total length of the meeting, $S$ is the set of the timestamps of all WRITE events. Note that we assume a final WRITE event at the end of the input.

Because our text data does not have timestamps, we cannot track processing latency. Instead, we track the \emph{redundancy factor} (RF) as a proxy metric. It is defined as the number of tokens sent to the summarization model divided by the number of tokens in the meeting. This metric gives an estimate of the extra cost of running a system in a real-world application over an offline one.

\subsection{Rewrites} 
For measuring rewriting, we use Normalized Erasure (NE, \citealt{arivazhagan2020re}) from simultaneous translation. This metric severely punishes changing a single word near the beginning of the output. Since we are examining only one system that rewrites its output, we leave a more detailed investigation of the user experience of rewrites in online summarization for future work.

\section{Evaluation}

\subsection{Final Summaries}
\label{sec:final}

\begin{table}
    \centering
    \begin{tabular}{lr|c|cr}
        \textbf{Model} & \textbf{$C$} & \textbf{R1} & \textbf{EL} & \textbf{RF}\\
        \hline
        \multicolumn{2}{l|}{\emph{Offline Baselines}}& & & \\
        GPT-4 & --- & 43.7 & --- & 1.0x\\
        Zoom-long & --- & 41.2 & --- & 1.0x\\
        \hline
        \multicolumn{2}{l|}{\emph{Length-Based}}& & & \\
        Bart & 768 & 42.7 & 25.9 & 1.0x\\
        GPT-4 & 1024 & 39.8 & 34.4 & 1.0x\\
        \hline
        \multicolumn{2}{l|}{\emph{Model-Based}}& & & \\
        Bart & 512 & 42.6 & 26.2 & 2.3x\\
        \hline
        \multicolumn{2}{l|}{\emph{Sliding Window}}& & & \\
        Bart & 256 & 41.8 & 11.7 & 4.1x\\
        GPT-4 & 1024 & 40.8 & 34.4 & 2.1x\\
        \hline
        \multicolumn{2}{l|}{\emph{Full Rewriting}}& & & \\
        GPT-4 & 512 & 41.5 & 17.2 & 12.2x \\
        \hline
        \multicolumn{2}{l|}{\emph{Full Incremental}} & & & \\
        GPT-4 & 2048 & 40.5 & 67.9 & 3.9x\\
        
    \end{tabular}
    \caption{System scores. For each policy and backend model, the system with the best final Rouge score is shown. For all results, see Appendix~\ref{app:curves}.}
    \label{tab:system_scores}
\end{table}

\paragraph{Automatic Metrics} Table~\ref{tab:system_scores} shows the scores of the different evaluated systems. We first examine quality and latency separately: Our highest scoring systems (e.\,g.\ \textit{length-based} Bart 768: R1 42.7) achieve comparable quality scores to the offline baselines, but there are significant differences in the Rouge scores of the different online systems (min 39.8, max 42.7), which clearly indicate that the different summarization units selected by the different policies affect the output quality. In terms of Rouge score, Bart systems outperform GPT-4 in the online setting, which may be due to the Samsum training data better matching the style of the reference summaries, boosting the Bart systems' Rouge scores. For latency, the EL scores support our intuition that systems operating on the same schedule have the same EL scores (there is a small difference due to tokenization).

When analyzing the behavior of our policies by cross-checking quality and latency, \textit{length-based} Bart 768 achieves the best Rouge score among our online systems (42.7), but the \textit{sliding window} system (41.8) remains competitive with it (for both backend summarizers, cf.\ GPT-4 with 39.8 vs.\ 40.8) and does so with lower latency (25.9 vs.\ 11.7 respectively), at the cost of greater processing redundancy (4.1x for Bart, 2.1x for GPT-4). In the \textit{sliding window} policy, the redundancy is directly tied to how often the model will amend its previous output. GPT-4 does so much more reliably than Bart, leading to lower redundancy. Perhaps by finetuning with carefully selected data, the Bart model could be adapted to continue its output more reliably.

The \textit{model-based} system also achieves a very good Rouge score (42.6), indicating that model-based approaches have promise, but its latency appears to make it uncompetitive with the \textit{length-based} Bart system. Interpreting the Expected Latency, we can confirm that the policy implements a non-trivial decision process: If the policy always selected the same number of chunks, its latency would be the same as a \textit{length-based} system with that chunk size. Instead, its latency is similar to the \textit{length-based} 768 system, meaning the policy uses both one- and two-chunk inputs. A stronger quality estimation model may further improve this system.

The \textit{full rewriting} policy scores well in terms of Rouge, but incurs a very high redundancy (and therefore cost), making it unattractive for most real applications, although a longer chunk size can reduce redundancy while increasing latency. It is the only rewriting system that we evaluated and while it did sometimes keep parts of the summary the same, it would often significantly change the text, reaching a Normalized Erasure of 13.8.

The \textit{fully incremental} policy does not achieve competitive scores with the other systems, mainly due to its very high latency. Because it also re-reads the entire meeting after each chunk, it also suffers from high redundancy.

\begin{table}
    \centering
    \begin{tabular}{r|ccc|c}
         \textbf{System} & \textbf{Ade} & \textbf{Flu} & \textbf{Rel} & \textbf{Ave} \\
         \hline
         Offline GPT-4 & 4.8 & 4.2 & 4.4 & 4.5 \\
         Offline Zoom-long & 4.2 & 3.8 & 4.2 & 4.0 \\
         \hline
         Length-Based Bart & 3.8 & 3.5 & 3.7 & 3.6\\
         Model-Based Bart & 4.6 & 4.1 & 4.5 & 4.4\\
         Sliding Window Bart & 4.2 & 3.8 & 4.2 & 4.1\\
         Rewriting GPT-4 & 3.9 & 4.3 & 4.3 & 4.2 \\
    \end{tabular}
    \caption{Human evaluation scores for final system summaries. Ade = Adequacy, Flu = Fluency, Rel = Relevance, Ave = Average}
    \label{tab:human_final}
\end{table}

\begin{figure*}
    \centering
    \includegraphics[width=\textwidth]{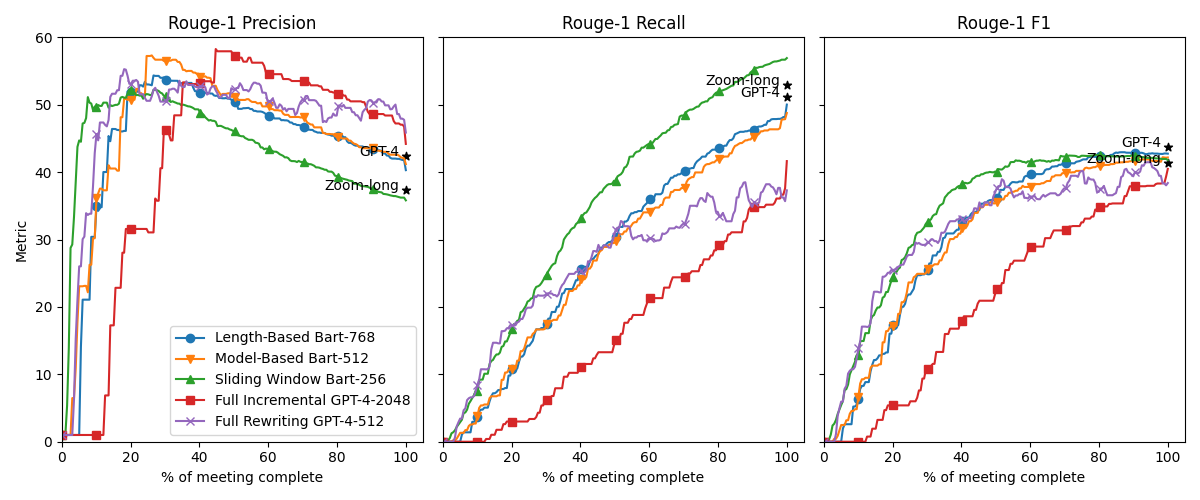}    
    \caption{Rouge curves for the best-performing system for each policy. All curves are in Appendix~\ref{app:curves}.}
    \label{fig:curves}
\end{figure*}

\begin{table*}
    \centering
    \begin{tabular}{r|ccc|c|c||c}
    \textbf{System} & \textbf{Ade} & \textbf{Flu} & \textbf{Rel} & \textbf{Ave} &  \textbf{R1 @ 40\%} & \textbf{R1-AUC} \\
    \hline
    Length-Based Bart-768 & 3.4 & 3.4 & 3.3 & 3.4 & 32.8 & 3047 \\
    Sliding Window Bart-256 & 3.8 & 3.6 & 3.6 & 3.7 & 38.2 & 3382 \\
    Model-Based Bart-512 & 3.5 & 3.5 & 3.4 & 3.5 & 32.7 & 3025 \\
    \end{tabular}
    \caption{Evaluation scores for 40\% of the meeting transcript. R1-AUC over the whole meeting.}
    \label{tab:human}
\end{table*}

\paragraph{Human Evaluation} It is notable that four online systems (\textit{sliding window}, \textit{model-based} and \textit{length-based} Bart, and \textit{full rewriting} GPT-4) score higher in terms of Rouge-1 than the Zoom-long offline baseline. This raises the question if online systems can generate better summaries than offline methods. We perform a human evaluation to answer this question:
We ask two annotators with experience in annotating the outputs of NLP tasks to evaluate these four systems, as well as the two offline baselines. The annotators are asked to rate the summaries according to adequacy, fluency and relevance on a 5-point Likert scale, 5 being the best (Table~\ref{tab:human_final}). More details about the human evaluation can be found in Appendix~\ref{app:human}.

GPT-4 is still rated highest, but the \textit{sliding window} Bart and \textit{rewriting} GPT-4 systems indeed match the (Bart-based) offline Zoom-long and the \textit{model-based} Bart system even exceeds its scores. Meanwhile, the \textit{length-based} system was judged much worse than its Rouge scores would suggest. These results indicate that dynamically choosing summarization units as opposed to a fixed-length policy substantially improves the output quality and that a policy which chooses units ideal \emph{for the specific summarizer} is at least as good as using the text semantics for topic segmentation (as Zoom-long does). More research is needed to determine whether the units selected by either the \textit{sliding window} or the \textit{model-based} algorithm constitute topic segments, or if they optimize some unseen quality that the model learns from the training data.

\paragraph{Summary} Based on our evaluations, the \textit{model-based} Bart system achieves the best quality among the online systems, while the \textit{sliding window} Bart system has better latency. Which system is more suitable will depend on the individual preference for quality vs.\ latency. In general, we find these results encouraging, indicating the strong performance of the online systems. However, we cannot fully evaluate the quality of these systems without examining the intermediate summaries.

\subsection{Intermediate Summaries}

\paragraph{Automatic Metrics} In summarization, latency is tightly connected to the quality of partial summaries. If a fact from the source appears earlier in a partial summary, it should affect both metrics for latency and partial summary quality. There exists no adequate measure for this relationship. We motivate a metric of our own with a visual inspection.

For a given model, we calculate a Rouge score every time there is an update to the output, always using the full human reference. We then interpret these discrete events as a continuous observation of Rouge score: At a given time, the observed Rouge score is the score of the summary as it is seen by the user at that time. Expressing the time as percent of the meeting completed allows us to average scores over a whole test set, yielding a more reliable measurement of Rouge over time.

Observing these curves (Figure \ref{fig:curves}) allows us to visually interpret the measured quality of intermediate summaries and inspect the different dynamics of the various systems throughout a meeting. We generally expect the recall score to increase over the course of the meeting and, ideally, the precision score to remain constant. In reality, we observe a drop in precision in almost all systems. Generally, the systems with longer outputs lose more precision over time than the more concise ones. The opposite is the case for recall: The systems with shorter outputs achieve better recall, not just in the final score but also in the intermediate summaries. This suggests that it is important to consider the F1 score, which balances precision and recall, for a better insight into the overall quality.

Observing the F1 graph reveals differences in the systems that the final score could not show: Even though most systems end up at a broadly similar score, they have appreciably different trajectories to arrive at that score. Depending on one's point of view, these trajectories can be interpreted as either a measure of latency (\textit{sliding window} Bart reaches a given F1 score earlier than \textit{length-based} Bart, even though both end at almost the same score) or quality (at any given time in the meeting, the summaries of \textit{sliding window} Bart rate higher than those of the \textit{length-based} one).
The \textit{model-based} system progresses on a very similar schedule as the \textit{length-based} one, which is reflected in the curves as well as the EL score.

\paragraph{Human Evaluation} We validate the expressiveness of the curves with a human evaluation, using the same setup as before. The only difference is that this time, the annotators are shown only the first 40\% of the meeting transcript and the summaries that are available at that time.
The results are shown in Table~\ref{tab:human}.
Their evaluation reinforces our confidence in dynamic segmentation strategies, with both dynamic systems outperforming the \textit{length-based} system.
The human annotators rate the \textit{model-based} and \textit{length-based} systems very similarly, with a slight preference toward the \textit{model-based system}. This mirrors the offline evaluations, where humans expressed a preference toward dynamic segmentation models, whereas the Rouge score fails to capture this slight difference. The significant preference of the annotators towards the sliding window is reflected by our metric, validating its expressiveness when there is a significant difference in perceived quality.

\paragraph{R1-AUC} Rendering and visually comparing curves for each system may not always be practical. We distill the information from the curves down to a single number by calculating the \emph{area under the curve} (AUC) for the Rouge-1 F1 curve (last column in Table~\ref{tab:human}). The convex shape of the \textit{sliding window} model's curve means that it scores very high on this metric (3382), whereas the more flat curve of the \textit{model-based} system scores lower (3025). Rouge curves and R1-AUC for all systems can be found in Appendix~\ref{app:curves}.

\paragraph{Summary} Online summarization is a relatively slow-paced task and the user will see intermediate summaries most of the time. It is therefore essential to compare systems not only on their final summary quality but also to take into account their performance over time. The human evaluation validates the expressiveness of our metric, R1-AUC, giving us a solid basis on which to make judgments about systems' intermediate summary quality. Based on these metrics, \textit{sliding window} Bart-256 is the system producing the best intermediate summaries.

\section{Conclusion and Future Work}

We conducted the first study of online meeting summarization, by proposing several policies and performing a detailed evaluation. We filled the gap of suitable metrics for latency and quality of intermediate summaries with our own novel metrics, expected latency and R1-AUC, respectively, allowing us to perform a multifaceted analysis of different policies to find the right trade-off between quality and latency. Our automatic and human evaluations showed that our proposed policies are able to create strong summaries of meetings in an online setting while being easy to deploy on top of an existing offline summarizer. They also showed the impact that policy choice has on both final and intermediate summary quality, with dynamic segmentation strategies outperforming static ones.

However, this initial paper also confirms that there is much room for further experimentation and innovation, both in the systems themselves and in methods of evaluation. Particularly promising directions are the development of learned policies (possibly observing from a human teacher), specialized architectures that can reuse more internal states to reduce redundancy and cost, and/or explicitly training with partial information like in simultaneous translation \citep{niehues2018low}.

We believe that the importance of this task is underappreciated in academic research and hope to inspire others to take up efforts in this direction.

\section*{Acknowledgment}

The work presented in this paper is partially funded by the European Union’s Horizon research and innovation program under grant agreement No 101135798, project Meetween (My Personal AI Mediator for Virtual MEETtings BetWEEN People).

\bibliography{anthology,custom}

\appendix

\clearpage

\section{Prompts and Hyperparameters}
\label{app:prompts}

We run all Bart experiments using beam search with a beam size of 4 and suppress recurring 3-grams. GPT-4 is run in sampling mode with a temperature of 0.7 with the following prompts:

\subsection{Length-Based}

\begin{mdframed}
\tt
Summarize the following project meeting in 2-3 bullet points:\\

\noindent John: Hello everyone.\\
...
\end{mdframed}

\subsection{Sliding Window}

Initial summary:
\begin{mdframed}
\tt
Summarize this conversation snippet in one sentence.\\
Reply only with the summary.\\

\noindent John: Hello everyone.\\
...
\end{mdframed}

Subsequent summaries:
\begin{mdframed}
\tt
More of the transcript has become available.\\
Either continue your previous summary with exactly one additional sentence, or reply with "No update necessary" if the new transcript does not warrant an update to the summary. Do not repeat what you wrote before.\\

\noindent Amy: As I was saying.\\
...
\end{mdframed}

\subsection{Full Rewriting}

\begin{mdframed}
\tt
Create a summary of the following meeting in N bullet points.\\

\noindent John: Hello everyone.\\
...
\end{mdframed}

The value of N is increased by one every chunk.

\subsection{Full Incremental}

Same as sliding window.

\section{Human Evaluation Details}
\label{app:human}

The annotators are [redacted for review copy] employees and are paid industry-standard wages. They are native English speakers with experience in NLP annotation tasks. Two annotators were asked to rate each transcript/summary pair with the following instructions:

\begin{mdframed}
The first column contains an incomplete transcript of a meeting. The transcript cuts off before the end, this is intentional.\\
The other columns contain automatic summaries created from that portion of the transcript, generated by different systems.\\
We would like you to rate the summaries according to three criteria: Adequacy, Fluency Relevance\\
\textbf{Adequacy} assesses if the summary adequately captures the major topics discussed in the meeting, also considering coverage (all such topics covered).\\
\textbf{Fluency} reflects if the summary consists of fluent, coherent text and is readable to the evaluator.\\
\textbf{Relevance} signifies the extent to which the summary overall captures the important content from the source transcript (as opposed to summarizing useless parts).\\
Please rate each summary from 1 (worst) to 5 (best).
\end{mdframed}

\newpage
\section{Pseudocode}
\label{app:pseudo}

\begin{algorithm}[H]
\begin{algorithmic}[1]
\Require A stream of chunks $S$, maximum input length \texttt{maxlen}
\Require \textsc{Summarize}$(t)$, \textsc{Rate}$(s)$
\While {Chunks remain in $S$}
\State \texttt{buffer} $\gets$ list($\{$next chunk in $S\}$)
\State \texttt{summaries}, \texttt{ratings} $\gets$ empty list
\Repeat
\State \texttt{s} $\gets$ \Call{Summarize}{\Call{Concat}{\texttt{buffer}}}
\State Append \texttt{s} to \texttt{summaries}
\State Append \Call{Rate}{s} to \texttt{ratings}
\State READ next chunk from $S$, append to \texttt{buffer}
\Until {$|$\Call{Concat}{\texttt{buffer}}$| >$ \texttt{maxlen} }
\State \texttt{i} $\gets $ argmax \texttt{ratings}
\State WRITE \texttt{summaries}[\texttt{i}]
\State Return \texttt{buffer}[$(\texttt{i}+1)$..] to $S$
\EndWhile
\end{algorithmic}
\caption{The model-based policy.}\label{alg:modelbased}
\end{algorithm}

\begin{algorithm}[H]
\begin{algorithmic}[1]
\Require A stream of chunks $S$
\Require \textsc{Summarize}$(t, p)$ where $p$ is a forced prefix
\State \texttt{prev\_in}, \texttt{prev\_out} $\gets \varnothing$
\State \texttt{cur\_in} $\gets \varnothing$
\While {Chunks remain in $S$}
\State READ next chunk from $S$, append to \texttt{cur\_in}
\State \texttt{input} $\gets$ \Call{Concat}{\texttt{prev\_in}, \texttt{cur\_in}}
\State \texttt{s} $\gets$ \Call{Summarize}{\texttt{input}, \texttt{prev\_in}}
\If {\texttt{s} $\ne$ \texttt{prev\_in}}
\State \texttt{cur\_out} $\gets$ \texttt{s}[$|$\texttt{prev\_out}$|$..]
\State WRITE \texttt{cur\_out}
\State \texttt{prev\_in}, \texttt{prev\_out} $\gets$ \texttt{cur\_in}, \texttt{cur\_out}
\State \texttt{cur\_in} $\gets \varnothing$
\EndIf
\EndWhile
\end{algorithmic}
\caption{The sliding window policy.}\label{alg:sliding}
\end{algorithm}

\newpage
\section{All System Scores}
\label{app:curves}

\begin{figure}[h]
    \centering
    \includegraphics[width=\columnwidth]{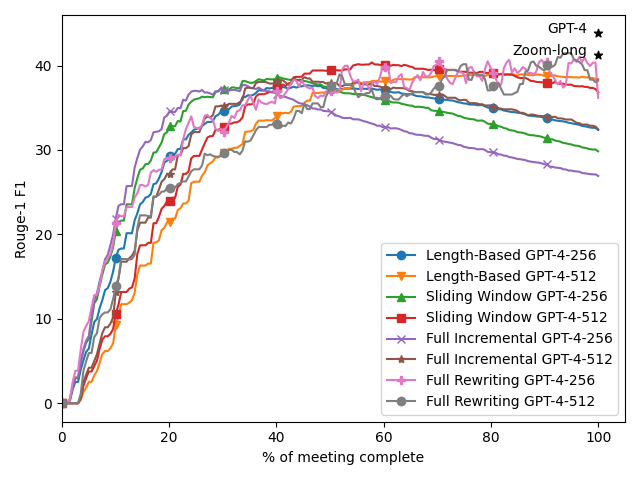}\\
    \includegraphics[width=\columnwidth]{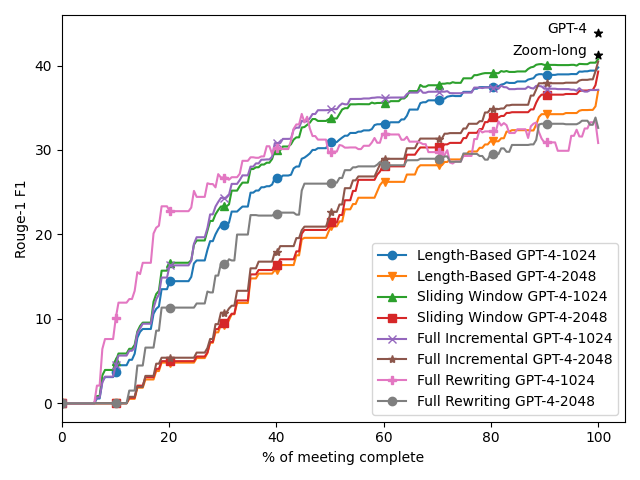}\\
    \includegraphics[width=\columnwidth]{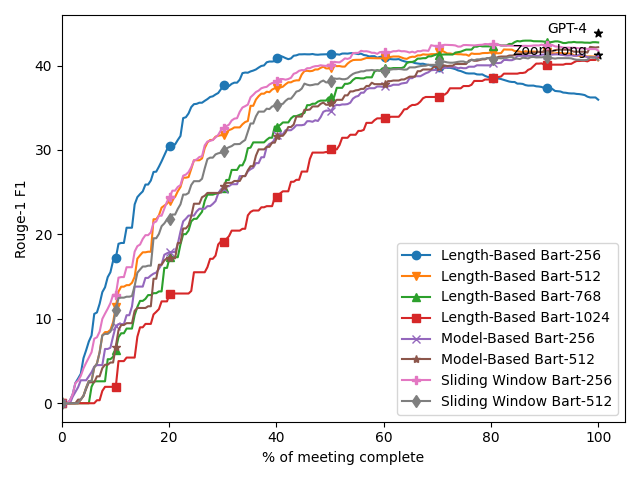}
    \caption{Rouge F1 curves for all systems}
\end{figure}

\begin{table*}
    \centering
    \begin{tabular}{lr|rrr|rrr|r|r}
    & & \multicolumn{3} {c|} {Quality} & \multicolumn{3} {c|} {Latency} & Rewriting & \makecell{Intermediate \\ Summaries}  \\ 
         \textbf{Model} & \textbf{$C$} & \textbf{R1} & \textbf{R2} & \textbf{Ratio} & \textbf{LAAL} & \textbf{EL} & \textbf{RF} & \textbf{NE} & \textbf{R1-AUC}\\
         \hline
         \multicolumn{2}{l|}{\emph{Offline}}& & & & & & & \\
         GPT-4 & --- & 43.7 & 10.5 & 1.1 & --- & ---& 1.0x & --- & ---\\
        Zoom-long & --- & 41.2 & 10.3 & 1.2 & --- & ---& 1.0x & --- & ---\\
         \hline
         \multicolumn{2}{l|}{\emph{Length-Based}}& & & & & & & \\
         Bart & 256 & 36.0 & 11.4 & 2.3 & 2.8 & 8.4 & 1.0x & 0 & 3427\\
         Bart & 512 & 40.6 & 11.5 & 1.5 & 24.1 & 17.1 & 1.0x & 0 & 3291\\
         Bart & 768 & 42.7 & 11.0 & 1.1 & 37.3 & 25.9 & 1.0x & 0 & 3047\\
         Bart & 1024 & 41.0 & 9.5 & 0.8 & 112.5 & 34.2 & 1.0x & 0 & 2575\\
         GPT-4 & 256 & 32.4 & 7.9 & 2.8 & 11.5 & 8.4 & 1.0x & 0 & 3133\\
         GPT-4 & 512 & 38.2 & 7.6 & 1.6 & 28.6 & 17.2 & 1.0x & 0 & 3044\\
         GPT-4 & 1024 & 39.8 & 6.9 & 1.0 & 77.5 & 34.4 & 1.0x & 0 & 2593\\
         GPT-4 & 2048 & 37.0 & 6.5 & 0.6 & 206.3 & 67.9 & 1.0x & 0 & 1862\\
         \hline
         \hline
         \multicolumn{2}{l|}{\emph{Model-Based}}& & & & & & & \\
         Bart & 256 & 41.0 & 11.9 & 1.6 & 13.8 & 22.8 & 4.8x & 0 & 3155\\
         Bart & 512 & 42.6 & 11.8 & 1.2 & 11.9 & 26.2 & 2.3x & 0 & 3025\\
         \hline
         \multicolumn{2}{l|}{\emph{Sliding Window}}& & & & & & & \\
         Bart & 256 & 41.8 & 12.0 & 1.4 & 2.5 & 11.7 & 4.1x & 0 & 3382\\
         Bart & 512 & 40.7 & 11.1 & 1.3 & 22.8 & 17.1 & 2.9x & 0 & 3166\\
         GPT-4 & 256 & 29.8 & 8.4 & 3.5 & 14.0 & 8.4 & 2.1x & 0 & 3157\\
         GPT-4 & 512 & 36.8 & 9.3 & 2.1 & 28.4 & 17.2 & 2.1x & 0 & 3194\\   
         GPT-4 & 1024 & 40.8 & 8.7 & 1.1 & 66.4 & 34.4 & 2.1x & 0 & 2785\\
         GPT-4 & 2048 & 39.3 & 6.9 & 0.6 & 189.1 & 67.9 & 2.1x & 0 & 1980\\
         \hline
         \multicolumn{2}{l|}{\emph{Full Rewriting}}& & & & & & & \\
         GPT-4 & 256 & 36.2 & 7.5 & 0.8 & 716.1 & 8.4 & 24.3x & 37.1 & 3340\\
         GPT-4 & 512 & 41.5 & 9.4 & 0.7 & 650.1 & 17.2 & 12.2x & 13.8 & 3125\\
         GPT-4 & 1024 & 37.3 & 7.1 & 0.5 & 501.8 & 34.4 & 6.6x & 7.1 & 2623\\
         GPT-4 & 2048 & 32.6 & 6.6 & 0.4 & 438.2 & 67.9 & 3.8x & 3.5 & 2115\\
         \hline
         \multicolumn{2}{l|}{\emph{Full Incremental}}& & & & & & & \\
         GPT-4 & 256 & 26.7 & 8.6 & 4.2 & -8.3 & 8.8 & 30.3x & 0 & 2961\\
         GPT-4 & 512 & 32.4 & 9.5 & 2.9 & 17.4 & 17.7 & 14.0x & 0 & 3106\\
         GPT-4 & 1024 & 37.1 & 8.5 & 1.7 & 74.5 & 34.4 & 7.4x & 0 & 2739\\
         GPT-4 & 2048 & 40.5 & 8.2 & 0.9 & 176.0 & 67.9 & 3.9x & 0 & 2060\\
    \end{tabular}
    \caption{System scores. \emph{Ratio} is the ratio of hypothesis length to reference length. LAAL is measured in dialog turns.}
    \label{tab:all_scores}
\end{table*}

\end{document}